\pdfoutput=1

\documentclass[11pt]{article}

\usepackage[]{acl}

\usepackage{times}
\usepackage{latexsym}

\usepackage[T1]{fontenc}

\usepackage[utf8]{inputenc}

\usepackage{microtype}

\usepackage{array}
\newcolumntype{P}[1]{>{\centering\arraybackslash}p{#1}}

\usepackage{trackchanges}
\addeditor{YJ}
\addeditor{GZ}
\addeditor{HC}

\newcommand{\hlc}[2][yellow]{{%
		\colorlet{foo}{#1}%
		\sethlcolor{foo}\hl{#2}}%
}

\usepackage{amsmath,amsfonts,amssymb,bm}
\usepackage{pifont} 
\usepackage{graphicx,subfigure,epsfig,fancybox} 
\usepackage{float}
\usepackage{color} 
\usepackage{multirow}
\usepackage{natbib}

\usepackage{tikz}
\usetikzlibrary{shapes.geometric, positioning, calc}
\usetikzlibrary{arrows,shapes,calc}
\usetikzlibrary{trees,positioning,mindmap,shadows,fit}
\usetikzlibrary{decorations.pathreplacing}
\usetikzlibrary{tikzmark} 
\usetikzlibrary{intersections} 

\newcommand{\revised}[1]{\textcolor{blue}{}}
\renewcommand{\vec}[1]{\boldsymbol{#1}}

\DeclareMathOperator*{\argmin}{argmin}
\DeclareMathOperator*{\argmax}{argmax}

\usepackage{adjustbox}
\usepackage{array}
\usepackage{booktabs}

\newcolumntype{R}[2]{%
    >{\adjustbox{angle=#1,lap=\width-(#2)}\bgroup}%
    l%
    <{\egroup}%
}

\usepackage{setspace}

\newcommand*{\affaddr}[1]{#1} 
\newcommand*{\affmark}[1][*]{\textsuperscript{#1}}
\newcommand*{\email}[1]{\texttt{#1}}

\title{Pathologies of Pre-trained Language Models in Few-shot Fine-tuning}

\author{%
	Hanjie Chen\affmark[1],\quad Guoqing Zheng\affmark[2],\quad Ahmed Hassan Awadallah\affmark[2],\quad Yangfeng Ji\affmark[1] \\
	\affaddr{\affmark[1]Department of Computer Science, University of Virginia, Charlottesville, VA, USA}\\
	\affaddr{\affmark[2]Microsoft Research}\\
	\email{\{hc9mx, yangfeng\}@virginia.edu}\\
	\email{\{zheng, hassanam\}@microsoft.com}\\
}

\begin{document}
\maketitle

\begin{abstract}
  Although adapting pre-trained language models with few examples has shown promising performance on text classification, there is a lack of understanding of where the performance gain comes from. 
In this work, we propose to answer this question by interpreting the adaptation behavior using post-hoc explanations from model predictions.
By modeling feature statistics of explanations, we discover that (1) without fine-tuning, pre-trained models (e.g. BERT and RoBERTa) show strong prediction bias across labels; (2) although few-shot fine-tuning can mitigate the prediction bias and demonstrate promising prediction performance, our analysis shows models gain performance improvement by capturing non-task-related features (e.g. stop words) or shallow data patterns (e.g. lexical overlaps). 
These observations alert that pursuing model performance with fewer examples may incur pathological prediction behavior, which requires further sanity check on model predictions and careful design in model evaluations in few-shot fine-tuning.
\end{abstract}

\section{Introduction}
\label{sec:intro}
Pre-trained language models \citep{NEURIPS2020_1457c0d6, liu2019roberta, devlin-etal-2019-bert} have shown impressive adaptation ability to dowstream tasks, achieving considerable performance even with scarce task-specific training data, i.e., few-shot adaptation \citep{radford2019language, schick-schutze-2021-exploiting, gao-etal-2021-making}. 
Existing  few-shot adaptation techniques broadly fall in fine-tuning and few-shot learning \citep{shin-etal-2020-autoprompt, schick-schutze-2021-just, chen-etal-2021-revisiting}. 
Specifically, fine-tuning includes directly tuning pre-trained language models with few task-specific examples or utilizing a natural-language prompt to transform downstream tasks to masked language modeling task for better mining knowledge from pre-trained models \citep{petroni-etal-2019-language, jiang-etal-2020-know, wang-etal-2021-transprompt}. 
Few-shot learning leverages unlabeled data or auxiliary tasks to provide additional information for facilitating model training \citep{zheng2021meta, wang2021meta, du-etal-2021-self}.

Although much success has been made in adapting pre-trained language models to dowstream tasks with few-shot examples, some issues have been reported. 
\citet{utama2021avoiding} found that models obtained from few-shot prompt-based fine-tuning utilize inference heuristics to make predictions on sentence pair classification tasks. 
\citet{zhao2021calibrate} discovered the instability of model performance towards different prompts in few-shot learning. 
These works mainly look at prompt-based fine-tuning and discover some problems.

This paper looks into direct fine-tuning and provides a different perspective on understanding model adaptation behavior via post-hoc explanations \citep{strumbelj2010efficient, sundararajan2017axiomatic}. 
Specifically, post-hoc explanations identify the important features (tokens) contribute to the model prediction per example. 
We model the statistics of important features over prediction labels via local mutual information (LMI) \citep{schuster-etal-2019-towards, du-etal-2021-towards}. 
We track the change of feature statistics with the model adapting from pre-trained to fine-tuned and compare it with the statistics of few-shot training examples. 
This provides insights on understanding model adaptation behavior and the effect of training data in few-shot settings.

We evaluate two pre-trained language models, BERT \citep{devlin-etal-2019-bert} and RoBERTa \citep{liu2019roberta}, on three tasks, including sentiment classification, natural language inference, and paraphrase identification. 
For each task, we test on both in-domain and out-of-domain datasets to evaluate the generalization of model adaptation performance.
We discover some interesting observations, some of which may have been overlooked in prior work: (1) without fine-tuning, pre-trained models show strong prediction bias across labels; (2) fine-tuning with a few examples can mitigate the prediction bias, but the model prediction behavior may be pathological by focusing on non-task-related features (e.g. stop words); (3) models adjust their prediction behaviors on different labels asynchronously; (4) models can capture the shallow patterns of training data to make predictions.
The insight drawn from the above observations is that pursuing model performance with fewer examples is dangerous and may cause pathologies in model prediction behavior. 
We argue that future research on few-shot fine-tuning or learning should do sanity check on model prediction behavior and ensure the performance gain is based on right reasons. 

\section{Setup }
\label{sec:setup}
\paragraph{Tasks.}
We consider three tasks: sentiment classification, natural language inference, and paraphrase identification. 
Each task contains an in-domain/out-of-domain dataset pair:  IMDB \citep{maas2011learning}/Yelp \citep{zhang2015character} for sentiment classification, SNLI \citep{bowman-etal-2015-large}/MNLI \citep{williams-etal-2018-broad} for natural language inference, and QQP \citep{iyer2017quora}/TwitterPPDB (TPPDB) \citep{lan-etal-2017-continuously} for paraphrase identification. 
The data statistics are in \autoref{tab:datasets} in Appendix \ref{sec:model_data}.

\paragraph{Models.}
We evaluate two pre-trained language models, BERT \citep{devlin-etal-2019-bert} and RoBERTa \citep{liu2019roberta}. 
For each task, we train the models on the in-domain training set with different ratio ($r \%, r \in [0, 1]$) of clean examples and then test them on in-domain and out-of-domain test sets. 

\paragraph{Explanations.}
We explain model prediction behavior via post-hoc explanations which identify important features (tokens) in input texts that contribute to model predictions. 
We test four explanation methods: sampling Shapley \citep{strumbelj2010efficient}, integrated gradients \citep{sundararajan2017axiomatic}, attentions \citep{mullenbach-etal-2018-explainable}, and individual word masks \citep{chen-etal-2021-explaining}. 
For each dataset, we randomly select 1000 test examples to generate explanations due to computational costs. 
We evaluate the faithfulness of these explanation methods via the AOPC metric \citep{nguyen-2018-comparing, chen-etal-2020-generating-hierarchical}. 
\autoref{tab:aopc} in Appendix \ref{sec:explanation} shows that the sampling Shapley generates more faithful explanations than other methods. 
In the following experiments, we adopt it to explain model predictions. 

More details about the models, datasets and explanations are in Appendix \ref{sec:sup_setup}.

\section{Experiments}
\label{sec:exp}
We report the prediction results (averaged across 5 runs) of BERT and RoBERTa trained with different ratio ($r \%: 0\sim1 \%$) of in-domain training examples on both in-domain and out-of-domain test sets in \autoref{tab:preds-s}. 
Overall, training with more examples, BERT and RoBERTa achieve better prediction accuracy on both in-domain and out-of-domain test sets.

We look into the predictions of models from pre-trained to fine-tuned and analyze model prediction behavior change during adaptation via post-hoc explanations. 
In \autoref{sec:exp_1}, we observe that pre-trained models without fine-tuning show strong prediction bias across labels. 
The models fine-tuned with a few examples can quickly mitigate the prediction bias by capturing non-task-related features, leading to a plausible performance gain. 
In \autoref{sec:exp_2}, we further quantify the prediction behavior change by comparing the feature statistics of model explanations and training data. We discover that the models adjust their prediction behavior on minority labels first rather than learning information from all classes synchronously and can capture the shallow patterns of training data, which may result in pathologies in predictions.

\begin{table*}[t]
	\small
	\centering
	\begin{tabular}{ccccccc}
		\toprule
		Models & IMDB & SNLI & QQP & Yelp & MNLI & TPPDB \\
		\midrule
		BERT & Pos & Neu & Pa & Pos & Neu & Pa \\
		\midrule
		RoBERTa & Pos & Con & Pa & Pos & Con & Pa \\
		\bottomrule
	\end{tabular}
	\caption{The majority labels of original pre-trained models on different datasets. Pos: postive, Con: contradiction, Neu: neutral, Pa: paraphrases.}
	\label{tab:major_label}
\end{table*}
\begin{table*}[t] 
	\small
	\centering
	\begin{tabular}{P{1.5cm}P{0.6cm}P{0.6cm}P{0.6cm}P{0.6cm}P{0.6cm}P{0.6cm}P{0.6cm}P{0.6cm}P{0.6cm}P{0.6cm}P{0.6cm}P{0.6cm}P{0.6cm}}
		\toprule
		& & \multicolumn{6}{c}{In-domain} & \multicolumn{6}{c}{Out-of-domain} \\
		\cmidrule(lr){3-8} \cmidrule(lr){9-14}
		Model & $r$ & \multicolumn{2}{c}{IMDB} & \multicolumn{2}{c}{SNLI} & \multicolumn{2}{c}{QQP} & \multicolumn{2}{c}{Yelp} & \multicolumn{2}{c}{MNLI} & \multicolumn{2}{c}{TPPDB} \\
		\cmidrule(lr){3-4} \cmidrule(lr){5-6}\cmidrule(lr){7-8} \cmidrule(lr){9-10}\cmidrule(lr){11-12} \cmidrule(lr){13-14}
		& & Acc & PB  & Acc & PB & Acc & PB & Acc & PB & Acc & PB & Acc & PB  \\
		\midrule
		\multirow{6}{*}{BERT}  &  0 & 49.73  &  \hlc[pink!97]{0.97}  & 35.30  & \hlc[pink!65]{0.65}  &  45.10 & \hlc[pink!46]{0.46}  & 49.86 & \hlc[pink!98]{0.98}  & 32.95 & \hlc[pink!95]{0.95}  & 44.44 & \hlc[pink!85]{0.85}  \\
		& 0.01 & -  & -  & 48.45 &  \hlc[pink!20]{0.20}  & 65.33 &  \hlc[pink!45]{0.45}  & - & - & 34.77 & \hlc[pink!92]{0.92}  & 80.25 &  \hlc[pink!35]{0.35}  \\
		& 0.05 & 60.31  & \hlc[pink!41]{0.41}   & 63.20 &  \hlc[pink!8]{0.08}  & 69.82 &  \hlc[pink!16]{0.16}  & 61.61 & \hlc[pink!9]{0.09}  & 37.58 & \hlc[pink!95]{0.95}  & 86.26 &   \hlc[pink!14]{0.14} \\
		& 0.1 &  70.76 &  \hlc[pink!13]{0.13}   & 69.13 & \hlc[pink!12]{0.12}   & 73.65 &  \hlc[pink!4]{0.04}  & 67.11 & \hlc[pink!41]{0.41}  & 38.27 & \hlc[pink!93]{0.93}  & 86.69 & \hlc[pink!7]{0.07}   \\
		& 0.5 & 84.71  &  \hlc[pink!5]{0.05}   & 77.63 &  \hlc[pink!6]{0.06}  & 79.06 &  \hlc[pink!2]{0.02}  & 88.19 & \hlc[pink!8]{0.08}  & 55.37 & \hlc[pink!45]{0.45}  & 87.27 & \hlc[pink!3]{0.03}   \\
		& 1 & 85.46  &   \hlc[pink!5]{0.05}  & 80.33 & \hlc[pink!6]{0.06}   & 80.16 &  \hlc[pink!5]{0.05}  & 89.09 & \hlc[pink!3]{0.03}  & 58.81 & \hlc[pink!34]{0.34}  & 85.22 & \hlc[pink!7]{0.07}   \\
		\midrule
		\multirow{6}{*}{RoBERTa} & 0  & 50.17  & \hlc[pink!100]{1.00}   &  33.55 & \hlc[pink!100]{1.00}  & 36.84  &  \hlc[pink!126]{1.26} & 50.00 & \hlc[pink!100]{1.00}  & 33.24 & \hlc[pink!102]{1.02}  & 18.93 & \hlc[pink!162]{1.62}  \\
		& 0.01 & -  &  -  & 36.27 & \hlc[pink!61]{0.61}   & 66.26 & \hlc[pink!54]{0.54}  & -  & - & 32.48 & \hlc[pink!100]{1.00}  & 81.07 &  \hlc[pink!38]{0.38}  \\
		& 0.05 & 58.11  &  \hlc[pink!61]{0.61}   & 68.03 & \hlc[pink!13]{0.13}   & 71.64 &  \hlc[pink!9]{0.09}  & 58.47 & \hlc[pink!71]{0.71}  & 42.41 &  \hlc[pink!88]{0.88}  & 82.30 &  \hlc[pink!21]{0.21}  \\
		& 0.1 & 78.58  &  \hlc[pink!10]{0.10}   & 77.04 &  \hlc[pink!7]{0.07}  & 76.82 &  \hlc[pink!4]{0.04}  & 76.59 & \hlc[pink!37]{0.37}  & 54.72 & \hlc[pink!75]{0.75}  & 83.54 &  \hlc[pink!21]{0.21}  \\
		& 0.5 &  89.56 &  \hlc[pink!1]{0.01}   & 83.84 &  \hlc[pink!4]{0.04}  & 81.91 &  \hlc[pink!5]{0.05}  & 92.54 & \hlc[pink!8]{0.08}  & 66.90 & \hlc[pink!37]{0.37}  & 85.67 &  \hlc[pink!6]{0.06}   \\
		& 1 & 90.34  &  \hlc[pink!1]{0.01}  & 85.43  & \hlc[pink!3]{0.03}  & 83.19  &  \hlc[pink!5]{0.05} & 93.76 & \hlc[pink!1]{0.01}  & 70.47 & \hlc[pink!20]{0.20}  & 85.78 & \hlc[pink!8]{0.08} \\
		\bottomrule
	\end{tabular}
	\caption{Prediction accuracy and bias of BERT and RoBERTa trained with different ratio ($r \%$) of in-domain training examples on both in-domain and out-of-domain test sets. Acc: accuracy (\%), PB: prediction bias. For PB, darker pink color implies larger prediction bias. Note that we do not consider $r=0.01$ for IMDB and Yelp datasets because the number of training examples is too small.}
	\label{tab:preds-s}
\end{table*}

\subsection{Prediction bias in pre-trained models}
\label{sec:exp_1}
In our pilot experiments, we find the predictions of pre-trained models without fine-tuning are biased across labels (see an example of confusion matrix in \autoref{fig:conf_mat} in Appendix \ref{sec:sup_exp}).
Original pre-trained models tend to predict all examples with a specific label on each dataset. 
We denote the specific label as the majority label and the rest labels as minority labels. 
The results of majority labels are in \autoref{tab:major_label}.

We propose a metric, prediction bias ($\text{PB}$), to quantify the bias of model predictions across labels,
\begin{eqnarray}
	\text{PB} = \left|\frac{T_{i_1} - T_{i_2}}{T_{i_1} + T_{i_2}} - \frac{D_{i_1} - D_{i_2}}{D_{i_1} + D_{i_2}} \right|, \label{eq:pred_bias_1} \\
	i_1 = \argmax\limits_{i\in \{1, \ldots, C\}}(T_{i}), i_2 = \argmin\limits_{i\in \{1, \ldots, C\}}(T_{i}) \nonumber
\end{eqnarray}
where $i_1$ and $i_2$ are the majority and most minority labels respectively. 
$T_i$ and $D_i$ denote the numbers of model predictions and test examples on label $i$ respectively, and $C$ is number of classes. 
The range of $\text{PB}$ is $[0, 2]$. 
PB takes $0$ if the label distribtion of model predictions is consistent with that of data. 
For balanced dataset, the upper bound of PB is $1$, that is all examples are predicted as one label. 
For imbalanced dataset, PB takes $2$ in an extreme case, where the dataset only contains one label of examples, while the model wrongly predicts them as another label. 
We consider data bias because some datasets (e.g. QQP and TPPDB) have imbalanced label distributions.

\begin{figure*}[t]
	\centering
	\includegraphics[width=0.75\textwidth]{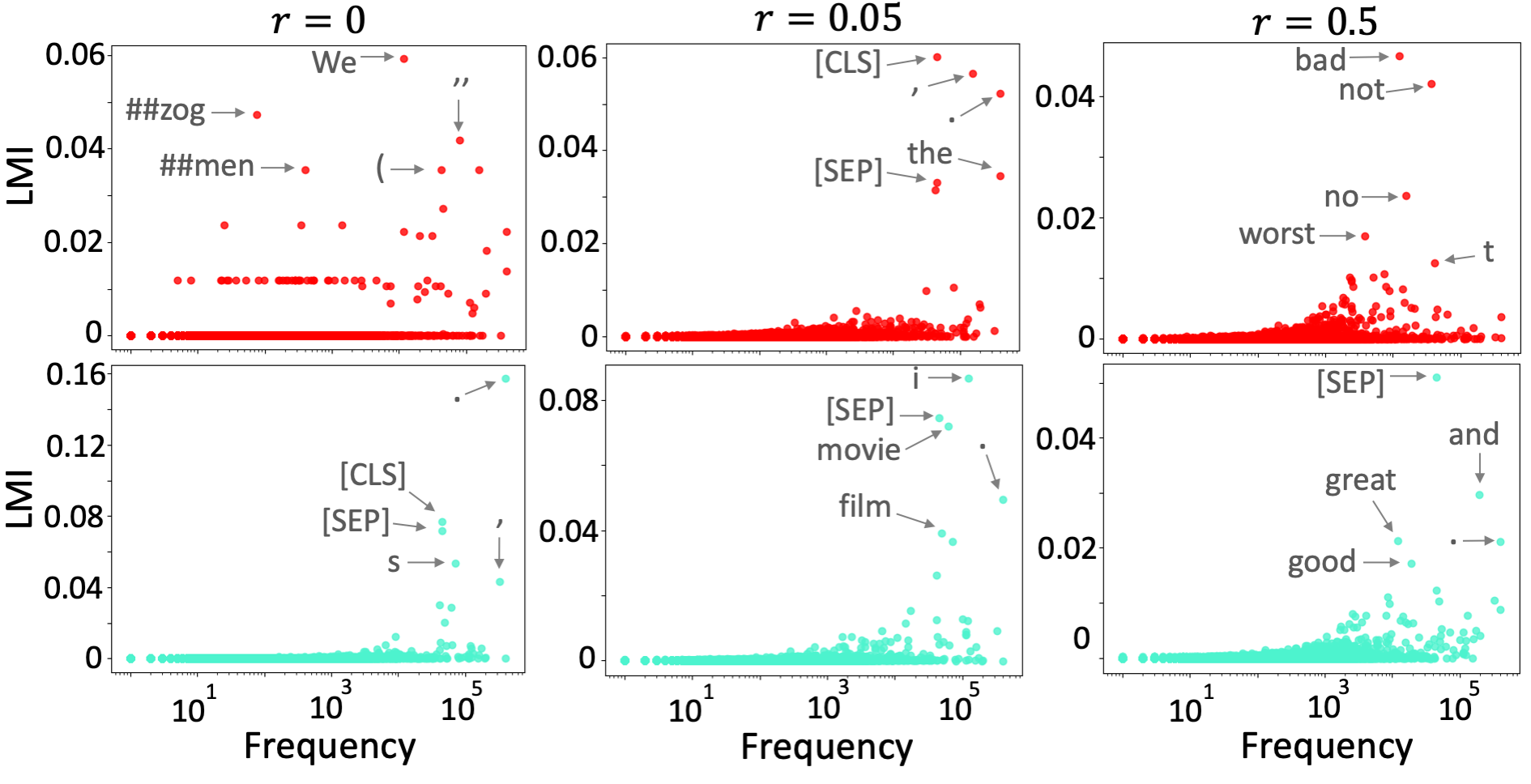}
	\caption{\label{fig:fea_dist} LMI distributions based on explanation statistics of BERT on the IMDB dataset with different $r$. The horizontal axis represents tokens in vocabulary in the ascending order of frequency. The upper and lower plots are on the negative and positive labels respectively. Top 5 tokens are pointed in each plot.}
\end{figure*}

\begin{table*}[t] 
	\tiny
	\centering
	\begin{tabular}{P{0.4cm}P{0.15cm}P{0.1cm}P{0.1cm}P{0.1cm}P{0.1cm}P{0.1cm}P{0.1cm}P{0.1cm}P{0.1cm}P{0.1cm}P{0.1cm}P{0.1cm}P{0.1cm}P{0.1cm}P{0.1cm}P{0.1cm}P{0.1cm}P{0.1cm}P{0.1cm}P{0.1cm}P{0.1cm}P{0.1cm}P{0.1cm}P{0.1cm}P{0.1cm}P{0.1cm}P{0.1cm}P{0.1cm}P{0.1cm}}
		\toprule
		& & \multicolumn{14}{c}{In-domain} & \multicolumn{14}{c}{Out-of-domain} \\
		\cmidrule(lr){3-16} \cmidrule(lr){17-30}
		\multirow{4}{*}{Model} & \multirow{4}{*}{$r$} & \multicolumn{4}{c}{IMDB} & \multicolumn{6}{c}{SNLI} & \multicolumn{4}{c}{QQP} & \multicolumn{4}{c}{Yelp} & \multicolumn{6}{c}{MNLI} & \multicolumn{4}{c}{TPPDB} \\
		\cmidrule(lr){3-6} \cmidrule(lr){7-12}\cmidrule(lr){13-16} \cmidrule(lr){17-20}\cmidrule(lr){21-26} \cmidrule(lr){27-30}
		& & \multicolumn{2}{c}{Ori}  & \multicolumn{2}{c}{Data}  & \multicolumn{3}{c}{Ori} & \multicolumn{3}{c}{Data} & \multicolumn{2}{c}{Ori} & \multicolumn{2}{c}{Data} & \multicolumn{2}{c}{Ori} & \multicolumn{2}{c}{Data} & \multicolumn{3}{c}{Ori} & \multicolumn{3}{c}{Data} & \multicolumn{2}{c}{Ori} & \multicolumn{2}{c}{Data}  \\
		\cmidrule(lr){3-4}\cmidrule(lr){5-6}\cmidrule(lr){7-9}\cmidrule(lr){10-12}\cmidrule(lr){13-14}\cmidrule(lr){15-16}\cmidrule(lr){17-18}\cmidrule(lr){19-20}\cmidrule(lr){21-23}\cmidrule(lr){24-26}\cmidrule(lr){27-28}\cmidrule(lr){29-30}
		& & Neg & Pos & Neg & Pos & En & Con & Neu & En & Con & Neu & NPa& Pa & NPa& Pa & Neg & Pos & Neg & Pos & En & Con & Neu & En & Con & Neu & NPa& Pa & NPa& Pa \\
		\midrule
		\multirow{6}{*}{BERT} & 0.01 & - & - & - & - & \hlc[pink!71]{0.71} & \hlc[pink!43]{0.43} & \hlc[pink!33]{0.33}  & \hlc[pink!70]{0.70} & \hlc[pink!42]{0.42} & \hlc[pink!51]{0.51}  & \hlc[pink!67]{0.67} & \hlc[pink!32]{0.32} & \hlc[pink!93]{0.93} & \hlc[pink!45]{0.45}  & - & - & -  & - & \hlc[pink!35]{0.35} & \hlc[pink!9]{0.09} & \hlc[pink!29]{0.29} & \hlc[pink!40]{0.40} & \hlc[pink!33]{0.33} & \hlc[pink!76]{0.76}  & \hlc[pink!74]{0.74} & \hlc[pink!16]{0.16} & \hlc[pink!155]{1.55} & \hlc[pink!18]{0.18}  \\
		& 0.05 & \hlc[pink!226]{2.26} & \hlc[pink!45]{0.45} & \hlc[pink!90]{0.90} & \hlc[pink!63]{0.63} & \hlc[pink!58]{0.58} & \hlc[pink!60]{0.60} & \hlc[pink!47]{0.47}  & \hlc[pink!31]{0.31} & \hlc[pink!17]{0.17} & \hlc[pink!16]{0.16}  & \hlc[pink!49]{0.49} & \hlc[pink!14]{0.14} & \hlc[pink!23]{0.23} & \hlc[pink!22]{0.22}  & \hlc[pink!220]{2.20} & \hlc[pink!69]{0.69} & \hlc[pink!66]{0.66} & \hlc[pink!43]{0.43}  & \hlc[pink!43]{0.43} & \hlc[pink!45]{0.45} & \hlc[pink!41]{0.41} &  \hlc[pink!76]{0.76} & \hlc[pink!27]{0.27} & \hlc[pink!63]{0.63} & \hlc[pink!87]{0.87} & \hlc[pink!2]{0.02} & \hlc[pink!58]{0.58} & \hlc[pink!3]{0.03}  \\
		& 0.1 & \hlc[pink!200]{2.00} & \hlc[pink!76]{0.76} & \hlc[pink!80]{0.80}  & \hlc[pink!54]{0.54} & \hlc[pink!56]{0.56} & \hlc[pink!82]{0.82} & \hlc[pink!45]{0.45}  & \hlc[pink!30]{0.30} & \hlc[pink!42]{0.42} & \hlc[pink!46]{0.46}  & \hlc[pink!46]{0.46} & \hlc[pink!53]{0.53} & \hlc[pink!19]{0.19} & \hlc[pink!37]{0.37}  & \hlc[pink!206]{2.06} & \hlc[pink!79]{0.79} & \hlc[pink!37]{0.37} & \hlc[pink!45]{0.45}  & \hlc[pink!46]{0.46} & \hlc[pink!49]{0.49} & \hlc[pink!41]{0.41} & \hlc[pink!61]{0.61} & \hlc[pink!40]{0.40} & \hlc[pink!125]{1.25}  & \hlc[pink!67]{0.67} & \hlc[pink!21]{0.21} & \hlc[pink!53]{0.53} & \hlc[pink!0]{0.00}  \\
		& 0.5 & \hlc[pink!139]{1.39} & \hlc[pink!80]{0.80} & \hlc[pink!116]{1.16} & \hlc[pink!52]{0.52} & \hlc[pink!70]{0.70} & \hlc[pink!151]{1.51} & \hlc[pink!94]{0.94}  & \hlc[pink!14]{0.14} & \hlc[pink!54]{0.54} & \hlc[pink!46]{0.46}  & \hlc[pink!31]{0.31} & \hlc[pink!67]{0.67} & \hlc[pink!8]{0.08} & \hlc[pink!21]{0.21} & \hlc[pink!161]{1.61} & \hlc[pink!93]{0.93} & \hlc[pink!73]{0.73} & \hlc[pink!52]{0.52}  & \hlc[pink!92]{0.92} & \hlc[pink!170]{1.70} & \hlc[pink!78]{0.78} &  \hlc[pink!82]{0.82} & \hlc[pink!91]{0.91} & \hlc[pink!103]{1.02} & \hlc[pink!93]{0.93} & \hlc[pink!9]{0.09} & \hlc[pink!37]{0.37} & \hlc[pink!4]{0.04}  \\
		& 1 & \hlc[pink!121]{1.21} & \hlc[pink!160]{1.60} & \hlc[pink!68]{0.68} & \hlc[pink!86]{0.86} & \hlc[pink!80]{0.80} & \hlc[pink!102]{1.02} & \hlc[pink!65]{0.65}  & \hlc[pink!14]{0.14} & \hlc[pink!48]{0.48} & \hlc[pink!52]{0.52}  & \hlc[pink!21]{0.21} & \hlc[pink!101]{1.01} & \hlc[pink!0]{0.00} & \hlc[pink!42]{0.42}  & \hlc[pink!73]{0.73} & \hlc[pink!194]{1.94} &  \hlc[pink!46]{0.46} & \hlc[pink!83]{0.83} & \hlc[pink!73]{0.73} & \hlc[pink!131]{1.31} & \hlc[pink!55]{0.55} & \hlc[pink!76]{0.76} & \hlc[pink!69]{0.69} & \hlc[pink!114]{1.14}  & \hlc[pink!46]{0.46} & \hlc[pink!54]{0.54} & \hlc[pink!33]{0.33} & \hlc[pink!11]{0.11}  \\
		\midrule
		\multirow{6}{*}{RoBERTa} & 0.01 & - & - & - & - & - & \hlc[pink!96]{0.96} & -  & \hlc[pink!76]{0.76} & \hlc[pink!52]{0.52} & \hlc[pink!56]{0.56}  & - & \hlc[pink!8]{0.08} & \hlc[pink!54]{0.54} & \hlc[pink!36]{0.36}  & - & - & - & - & - & \hlc[pink!95]{0.95} & - & \hlc[pink!33]{0.33} & \hlc[pink!84]{0.84} & \hlc[pink!95]{0.95}  & - & 0.00 &  \hlc[pink!155]{1.55} & 0.00 \\
		& 0.05 & - & \hlc[pink!66]{0.66} & \hlc[pink!17]{0.17} & \hlc[pink!72]{0.72} & - & \hlc[pink!62]{0.62} & -  & \hlc[pink!50]{0.50} & \hlc[pink!32]{0.32} & \hlc[pink!67]{0.67}  & - & \hlc[pink!43]{0.43} & \hlc[pink!22]{0.22} & \hlc[pink!35]{0.35}  & -  & \hlc[pink!38]{0.38} & \hlc[pink!14]{0.14}  & \hlc[pink!62]{0.62}  & - & \hlc[pink!26]{0.26} & - & \hlc[pink!89]{0.89} & \hlc[pink!22]{0.22} & \hlc[pink!107]{1.07}  & - & \hlc[pink!26]{0.26} & \hlc[pink!143]{1.43} & \hlc[pink!39]{0.39}  \\
		& 0.1 & - & \hlc[pink!103]{1.03} & \hlc[pink!69]{0.69} & \hlc[pink!71]{0.71} & - & \hlc[pink!105]{1.05} & - & \hlc[pink!22]{0.22} & \hlc[pink!57]{0.57} & \hlc[pink!45]{0.45}  & - & \hlc[pink!127]{1.27} & \hlc[pink!17]{0.17} & \hlc[pink!59]{0.59}  & - & \hlc[pink!96]{0.96}  & \hlc[pink!30]{0.30}  & \hlc[pink!47]{0.47}  & - & \hlc[pink!18]{0.18} & - & \hlc[pink!105]{1.05} & \hlc[pink!10]{0.10}  & \hlc[pink!62]{0.62}  & - & \hlc[pink!39]{0.39} & \hlc[pink!72]{0.72} & \hlc[pink!36]{0.36}  \\
		& 0.5 & - & \hlc[pink!133]{1.33} & \hlc[pink!81]{0.81} & \hlc[pink!42]{0.42} & - & \hlc[pink!207]{2.07} & -  & \hlc[pink!21]{0.21} & \hlc[pink!60]{0.60} & \hlc[pink!55]{0.55}  & - & \hlc[pink!101]{1.01} & \hlc[pink!15]{0.15} & \hlc[pink!69]{0.69}  & - & \hlc[pink!170]{1.70} & \hlc[pink!66]{0.66} & \hlc[pink!43]{0.43}  & - & \hlc[pink!70]{0.70} & - & \hlc[pink!87]{0.87} & \hlc[pink!70]{0.70} & \hlc[pink!79]{0.79}  & - & \hlc[pink!59]{0.59}  & \hlc[pink!79]{0.79} & \hlc[pink!48]{0.48}  \\
		& 1 & - & \hlc[pink!141]{1.41} & \hlc[pink!86]{0.86} & \hlc[pink!62]{0.62} & - & \hlc[pink!30]{0.30} & -  & \hlc[pink!17]{0.17} & \hlc[pink!32]{0.32} & \hlc[pink!23]{0.23}  & - & \hlc[pink!42]{0.42} & \hlc[pink!27]{0.27} & \hlc[pink!23]{0.23}  & - & \hlc[pink!191]{1.91} & \hlc[pink!65]{0.65}  & \hlc[pink!78]{0.78} & - & \hlc[pink!18]{0.18} & - & \hlc[pink!72]{0.72} & \hlc[pink!66]{0.66} & \hlc[pink!51]{0.51}  & - & \hlc[pink!64]{0.64} & \hlc[pink!95]{0.95} & \hlc[pink!47]{0.47}  \\
		\bottomrule
	\end{tabular}
	\caption{The KL divergence between LMI distributions. The columns of ``Ori'' and ``Data'' show the results with original pre-trained models' explanations or few-shot training data as the reference respectively. Neg: negative, Pos: postive, En: entailment, Con: contradiction, Neu: neutral, NPa: nonparaphrases, Pa: paraphrases. Darker color indicates larger KL divergence.}
	\label{tab:exp_kl_s}
\end{table*}

The results in \autoref{tab:preds-s} show that both pre-trained BERT and RoBERTa have strong prediction bias on all of the datasets.
The prediction bias decreases with models fine-tuned with more examples. 

\paragraph{Models make biased predictions by focusing on non-task-related features.}
To understand which features are associated with model prediction labels, we follow \citet{schuster-etal-2019-towards, du-etal-2021-towards} and analyze the statistics of model explanations via local mutual information (LMI). 
Specifically, we select top $k$ important features in each explanation and get a set of important features ($E=\{e\}$) over all explanations. 
We empirically take $k=10$ for the IMDB and Yelp datasets and $k=6$ for other datasets based on their average sentence lengths. 
The LMI between a feature $e$ and a particular label $y$ is 
\begin{equation}
	\label{eq:lmi}
	\text{LMI}(e, y)=p(e, y)\cdot \log \left(\frac{p(y \mid e)}{p(y)}\right),
\end{equation}
where $p(y \mid e)=\frac{count(e, y)}{count(e)}$, $p(y)=\frac{count(y)}{\lvert E \rvert}$, $p(e, y)=\frac{count(e, y)}{\lvert E \rvert}$, and $\lvert E \rvert$ is the number of occurrences of all features in $E$. 
Then we can get a distribution of LMI over all tokens in the vocabulary ($\{w\}$) built upon the dataset, i.e.
\begin{equation}
	\label{eq:p_lmi}
	P_{\text{LMI}}(w, y) = 
	\begin{cases}
		\text{LMI}(w, y) & \text{if token $w \in E$}\\
		0 & \text{else}
	\end{cases}
\end{equation}
We normalize the LMI distribution by dividing each value with the sum of all values. 

\autoref{fig:fea_dist} shows LMI distributions of BERT on the IMDB dataset with different $r$, where top 5 tokens are pointed in each plot (see \autoref{tab:top_fea} in \autoref{sec:sup_exp} for more results on other datasets). 
When $r=0$, we can see that BERT makes biased predictions on the positive label (in \autoref{tab:major_label}) by focusing on some non-task-related high-frequency tokens. 
The top features associated with the negative label include some relatively low-frequency tokens (e.g. \#\#men, \#\#zog) which may have been seen by the model during pre-training. 

\paragraph{Models adjust prediction bias by capturing non-task-related features on minority labels.}
Fine-tuning BERT with a few examples ($r=0.05$, exactly $9$ examples) from IMDB can quickly mitigate the prediction bias along with a plausible improvement on prediction accuracy (in \autoref{tab:preds-s}). 
However, \autoref{fig:fea_dist} (the middle upper plot) shows that the model captures non-task-related high-frequency tokens to make predictions on the minority label (negative), implying the performance gain is not reasonable. 
Only when the model is fine-tuned with more examples ($r=0.5$), it starts capturing task-specific informative tokens, such as ``bad'', ``good''.

\subsection{Quantifying model adaptation behavior}
\label{sec:exp_2}
To quantify the model prediction behavior change (in \autoref{fig:fea_dist}) during adaptation, we compute the Kullback–Leibler divergence (KLD) between the LMI distributions of the model without/with fine-tuning, i.e. $KL_{y}(P_{LMI}^{0}(w, y), P_{LMI}^{r}(w, y))$. 
The superscripts (``$0$'' or ``$r$'') indicate the ratio of training examples used in fine-tuning. 
Besides, we also evaluate how much the model prediction behavior is learned from the patterns of training data. 
Specifically, we compute the LMI distribution of few-shot training examples via \autoref{eq:lmi} and \autoref{eq:p_lmi}, except that $E$ represents the set of features appearing in those examples. 
Then we use the LMI distribution of data as the reference and compute the KLD between it and the LMI distribution of model explanations.

\autoref{tab:exp_kl_s} records the results of KLD with the LMI distribution of original pre-trained model explanations as the reference (columns of ``Ori'') or that of training data as the reference (columns of ``Data''). 
Note that we do not have the results of RoBERTa on some labels (e.g. ``Neg'') in ``Ori'' columns because the pre-trained RoBERTa does not make any predictions on those labels and we do not have the reference LMI distributions. 

\paragraph{Models adjust their prediction behaviors on different labels asynchronously.}
In ``Ori'' columns, the KLDs on minority labels are larger than those on majority labels when $r$ is small (e.g. 0.05). The changes of KLDs are discrepant across labels with $r$ increasing. The results show that the models focus on adjusting their prediction behavior on minority labels first rather than learning from all classes synchronously in few-shot settings.   

\paragraph{Models can capture the shallow patterns of training data.}
In ``Data'' columns, the KLDs on SNLI and QQP are overall smaller than those on IMDB, illustrating that it is easier for models to learn the patterns of datasets on sentence-pair classification tasks. 
With $r$ increasing, the KLDs on the entailment label of SNLI are smaller than those on other labels, which validates the observations in previous work \citep{utama2021avoiding, Nie_Wang_Bansal_2019} that models can capture lexical overlaps to predict the entailment label. 
Another interesting observation is the KLDs on Yelp in ``Data'' columns are mostly smaller than those on IMDB. 
This indicates that models may rely on the shallow patterns of in-domain datasets to make predictions on out-of-domain datasets. 

\section{Conclusion}
\label{sec:conclusion}
In this work, we take a closer look into the adaptation behavior of pre-trained language models in few-shot fine-tuning via post-hoc explanations. 
We discover many pathologies in model prediction behavior. 
The insight drawn from our observations is that promising model performance gain in few-shot learning could be misleading. 
Future research on few-shot fine-tuning or learning requires sanity check on model prediction behavior and some careful design in model evaluation and analysis.

\section*{Acknowledgments}
We thank the anonymous reviewers for many valuable comments.

\bibliography{custom}
\bibliographystyle{acl_natbib}

\clearpage
\newpage
\appendix
\section{Supplement of Setup}
\label{sec:sup_setup}

\subsection{Models and Datasets}
\label{sec:model_data}
We adopt the pretrained BERT-base and RoBERTa-base models from Hugging Face\footnote{\url{https://github.com/huggingface/pytorch-transformers}{}}. 
For sentiment classification, we utilize movie reviews IMDB \citep{maas2011learning} as the in-domain dataset and Yelp reviews \citep{zhang2015character} as the out-of-domain dataset. 
For natural language inference, the task is to predict the semantic relationship between a premise and a hypothesis as entailment, contradiction, or neutral. 
The Stanford Natural Language Inference (SNLI) corpus \citep{bowman-etal-2015-large} and Multi-Genre Natural Language Inference (MNLI) \citep{williams-etal-2018-broad} are used as the in-domain and out-of-domain datasets respectively. 
The task of paraphrase identification is to judge whether two input texts are semantically equivalent or not. 
We adopt the Quora Question Pairs (QQP) \citep{iyer2017quora} as the in-domain dataset, while using the TwitterPPDB (TPPDB) \citep{lan-etal-2017-continuously} as the out-of-domain dataset. 
\autoref{tab:datasets} shows the statistics of the datasets. 
 \begin{table*}[t] 
 	\centering
 	\small
 	\begin{tabular}{ccccccc}
 		\toprule
 		Datasets & \textit{C} & \textit{L} & \textit{\#train} & \textit{\#dev} & \textit{\#test} & Label distribution \\
 		\midrule
 		IMDB & 2 & 268 & 19992 & 4997 & 24986 & Positive: \textit{train}(10036), \textit{dev}(2414), \textit{test}(12535) \\
 		 &  &  &  &  &  & Negative: \textit{train}(9956), \textit{dev}(2583), \textit{test}(12451) \\
 		 \midrule
 		Yelp & 2 & 138 & 500000 & 60000 & 38000 & Positive: \textit{train}(250169), \textit{dev}(29831), \textit{test}(19000) \\
 		 &  &  &  &  &  & Negative: \textit{train}(249831), \textit{dev}(30169), \textit{test}(19000) \\
 		 \midrule
 		SNLI & 3 & 14 & 549367 & 4921 & 4921 &  Entailment: \textit{train}(183416), \textit{dev}(1680), \textit{test}(1649) \\
 		 &  &  &  &  &  & Contradiction: \textit{train}(183187), \textit{dev}(1627), \textit{test}(1651) \\
 		 &  &  &  &  &  & Neutral: \textit{train}(182764), \textit{dev}(1614), \textit{test}(1651) \\
 		 \midrule
 		MNLI & 3 & 22 & 391176 & 4772 & 4907 & Entailment: \textit{train}(130416), \textit{dev}(1736), \textit{test}(1695) \\
 		 &  &  &  &  &  & Contradiction: \textit{train}(130381), \textit{dev}(1535), \textit{test}(1631) \\
 		 &  &  &  &  &  & Neutral: \textit{train}(130379), \textit{dev}(1501), \textit{test}(1581) \\
 		 \midrule
 		QQP & 2 & 11 & 363178 & 20207 & 20215 & Paraphrases: \textit{train}(134141), \textit{dev}(7435), \textit{test}(7447) \\
 		 &  &  &  &  &  & Nonparaphrases: \textit{train}(229037), \textit{dev}(12772), \textit{test}(12768) \\
 		 \midrule
 		TPPDB & 2 & 15 & 42200 & 4685 & 4649 & Paraphrases: \textit{train}(11167), \textit{dev}(941), \textit{test}(880) \\
 		 &  &  &  &  &  & Nonparaphrases: \textit{train}(31033), \textit{dev}(3744), \textit{test}(3769) \\
 		\bottomrule
 	\end{tabular}
 	\caption{Summary statistics of the datasets, where \textit{C} is the number of classes, \textit{L} is average sentence length, and \textit{\#} counts the number of examples in the \textit{train/dev/test} sets. For label distribution, the number of examples with the same label in \textit{train/dev/test} is noted in bracket.}
 	\label{tab:datasets}
 \end{table*}

We implement the models in PyTorch 3.6. We set hyperparameters as: learning rate is $1e-5$, maximum sequence length is $256$, maximum gradient norm is $1$, and batch size is $8$. All experiments were performed on a single NVidia GTX 1080 GPU. We report the time for training each model on each in-domain dataset (with full training examples) in \autoref{tab:run_time}. 

\begin{table*}[tbh]
	\small
	\centering
	\begin{tabular}{cccc}
		\toprule
		Models & IMDB & SNLI & QQP  \\
		\midrule
		BERT & 856.43 & 25402.52 & 17452.12 \\
		\midrule
		RoBERTa & 912.47 & 256513.98 & 17514.80  \\
		\bottomrule
	\end{tabular}
	\caption{The average runtime (s/epoch) of each model on each in-domain dataset.}
	\label{tab:run_time}
\end{table*}

\subsection{Explanations}
\label{sec:explanation}
We adopt four explanation methods:
\begin{itemize}
	\item sampling Shapley (SS) \citep{strumbelj2010efficient}: computing feature attributions via sampling-based Shapley value \citep{shapley1953value}; 
	\item integrated gradients (IG) \citep{sundararajan2017axiomatic}: computing feature attributions by integrating gradients of points along a path from a baseline to the input; 
	\item attentions (Attn) \citep{mullenbach-etal-2018-explainable}: attention weights in the last hidden layer as feature attributions; 
	\item individual word masks (IMASK) \citep{chen-etal-2021-explaining}: learning feature attributions via variational word masks \citep{chen-ji-2020-learning}.
\end{itemize}

\paragraph{Explanation faithfulness.}
An important criterion for evaluating explanations is their faithfulness to model predictions \citep{jacovi-goldberg-2020-towards}. 
We evaluate the faithfulness of the four explanation methods via the AOPC metric \citep{nguyen-2018-comparing, chen-etal-2020-generating-hierarchical}. 
AOPC calculates the average change of prediction probability on the predicted class over all examples by removing top $1\ldots u$ words identified by explanations.
\begin{equation}
	\label{eq:aopc}
	\text{AOPC}=\frac{1}{U+1}\langle\sum_{u=1}^Up(y|\vec{x})-p(y|\vec{x}_{\backslash1\ldots u})\rangle_{\vec{x}},
\end{equation}
where $p(y|\vec{x}_{\backslash1\ldots u})$ is the probability for the predicted class when words $1\ldots u$ are removed and $\langle\cdot \rangle_{\vec{x}}$ denotes the average over all test examples. 
Higher AOPC score indicates better explanations. 

We test the BERT and RoBERTa trained with $1 \%$ in-domain training examples on each task.
For each dataset, we randomly select 1000 test examples to generate explanations due to computational costs. 
We report the results of AOPC scores when U = 10 in \autoref{tab:aopc}.
Sampling Shapley consistently outperforms other three explanation methods in explaining different models on both in-domain and out-of-domain datasets. 

\begin{table*}[t] 
	\small
	\centering
	\begin{tabular}{P{1.0cm}P{0.8cm}P{0.8cm}P{0.8cm}P{0.8cm}P{0.8cm}P{0.8cm}P{0.8cm}P{0.8cm}P{0.8cm}P{0.8cm}}
		\toprule
		& & \multicolumn{3}{c}{In-domain} & \multicolumn{3}{c}{Out-of-domain} \\
		\cmidrule(lr){3-5} \cmidrule(lr){6-8}
		Model & $r$ & IMDB & SNLI & QQP & Yelp &MNLI & TPPDB \\
		\midrule
		\multirow{4}{*}{BERT}  & SS & 0.41 & 0.82 & 0.61 & 0.53 & 0.77 &  0.40  \\
		& IG & 0.08 & 0.34 & 0.19 & 0.12 & 0.31 & 0.10   \\
		& Attn & 0.07 & 0.35 & 0.28 & 0.12 & 0.26 &  0.14  \\
		& IMASK & 0.09 & 0.28 & 0.25 & 0.09 & 0.25 & 0.08   \\
		\midrule
		\multirow{4}{*}{RoBERTa} & SS  & 0.25 & 0.86 & 0.53 & 0.28 & 0.84 &  0.28  \\
		& IG & 0.02 & 0.36 & 0.21 & 0.04 & 0.38 & 0.09   \\
		& Attn & 0.02 & 0.33 & 0.26 & 0.03 & 0.23 &  0.09  \\
		& IMASK & 0.02 & 0.18 & 0.18 & 0.03 & 0.17 &  0.05  \\
		\bottomrule
	\end{tabular}
	\caption{AOPC scores of different explanation methods in explaining different models.}
	\label{tab:aopc}
\end{table*}

\section{Supplement of Experiments}
\label{sec:sup_exp}

\begin{figure}[t]
	\centering
	\includegraphics[width=0.35\textwidth]{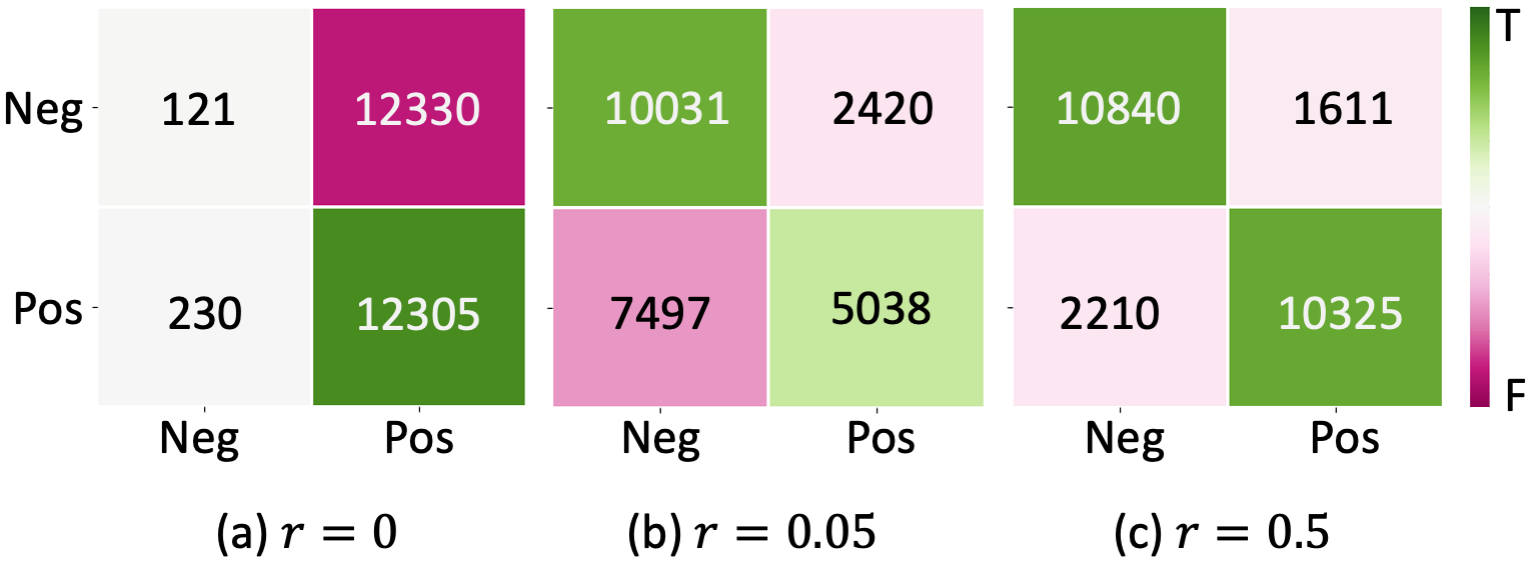}
	\caption{\label{fig:conf_mat} Confusion matrix of BERT (with different $r$) on the IMDB dataset. ``Neg'' and ``Pos'' represent negative and positive labels respectively. Vertical and horizontal dimensions show ground-truth and predicted labels respectively. Green and pink colors represent true or false predictions respectively. Darker color indicates larger number.}
\end{figure}

\begin{table*}[tbph]
	\small
	\centering
	\begin{tabular}{P{1cm}P{1cm}P{1cm}P{8cm}}
		\toprule
		Datasets & $r$ & Labels & Top Features\\
		\midrule
		\multirow{4}{*}{IMDB} & \multirow{2}{*}{0} & Neg & we \#\#zog " \#\#men ( ' [SEP] capitalism lynch hell \\
		\cmidrule(lr){3-4}
		& & Pos & . [CLS] [SEP] s , t movie film plot ) \\
		\cmidrule(lr){2-4}
		& \multirow{2}{*}{0.5} & Neg & bad not no worst t off terrible nothing stupid boring  \\
		\cmidrule(lr){3-4}
		& & Pos & [SEP] and great . good [CLS] love , film characters\\
		\cmidrule(lr){1-4}
		\multirow{4}{*}{Yelp} & \multirow{2}{*}{0} & Neg & . they majestic adds state owners loud dirty priced thai \\
		\cmidrule(lr){3-4}
		& & Pos & . [CLS] [SEP] , s t for i you m \\
		\cmidrule(lr){2-4}
		& \multirow{2}{*}{0.5} & Neg & not no bad t worst never off rude over nothing \\
		\cmidrule(lr){3-4}
		& & Pos & [SEP] great and good . [CLS] amazing love friendly experience \\
		\cmidrule(lr){1-4}
		\multirow{6}{*}{SNLI} & \multirow{3}{*}{0} & En & a [SEP] man the woman dog sitting sits his fire  \\
		\cmidrule(lr){3-4}
		& & Con & [SEP] [CLS] is the a , are in of there \\
		\cmidrule(lr){3-4}
		& & Neu & . people woman girl are playing looking [CLS] group boy \\
		\cmidrule(lr){2-4}
		& \multirow{3}{*}{0.5} & En & [SEP] . [CLS] and is a man there woman people \\
		\cmidrule(lr){3-4}
		& & Con & the a in [SEP] at sitting with man on playing \\
		\cmidrule(lr){3-4}
		& & Neu & [SEP] are for . man [CLS] is the a girl \\
		\cmidrule(lr){1-4}
		\multirow{6}{*}{MNLI} & \multirow{3}{*}{0} & En & the [SEP] \#\#ists israel ' recession ata consultants discusses attacked \\
		\cmidrule(lr){3-4}
		& & Con & [SEP] [CLS] , s to of in . the not \\
		\cmidrule(lr){3-4}
		& & Neu & . [CLS] they we you people about it really i \\
		\cmidrule(lr){2-4}
		& \multirow{3}{*}{0.5} & En & . [CLS] and is [SEP] there are , was of  \\
		\cmidrule(lr){3-4}
		& & Con & the ' . not no t [CLS] don to didn \\
		\cmidrule(lr){3-4}
		& & Neu & [SEP] [CLS] the for to all when . you it  \\
		\cmidrule(lr){1-4}
		\multirow{4}{*}{QQP} & \multirow{3}{*}{0} & NPa & ? is the a ' what india does quo why \\
		\cmidrule(lr){3-4}
		& & Pa & [SEP] [CLS] ? in i , of . best s \\
		\cmidrule(lr){2-4}
		& \multirow{2}{*}{0.5} & NPa & ? what [CLS] is how , why a the .\\
		\cmidrule(lr){3-4}
		& & Pa & [SEP] quo [CLS] best trump \#\#ra india life your sex \\
		\cmidrule(lr){1-4}
		\multirow{4}{*}{TPPDB} & \multirow{2}{*}{0} & NPa & trump ' the obama " we is russia a says \\
		\cmidrule(lr){3-4}
		& & Pa & [SEP] . [CLS] , s of in to \#\#t t \\
		\cmidrule(lr){2-4}
		& \multirow{2}{*}{0.5} & NPa & . , [CLS] ? '\textcircled{a} ; - a is \\
		\cmidrule(lr){3-4}
		& & Pa & [SEP] trump [CLS] inauguration obama russia repeal \#\#care cia senate \\
		\bottomrule
	\end{tabular}
	\caption{Top 10 important tokens for BERT predictions on different labels. Neg: negative, Pos: postive, En: entailment, Con: contradiction, Neu: neutral, NPa: nonparaphrases, Pa: paraphrases.}
	\label{tab:top_fea}
\end{table*}

\end{document}